\newif\ifcustomclass
\customclasstrue
\ifcustomclass
    \documentclass[10pt, a4paper, logo, twocolumn]{kiwi}
\else
    \documentclass[letterpaper, 10 pt, conference]{ieeeconf}
    \IEEEoverridecommandlockouts
    \usepackage[T1]{fontenc}
    \usepackage{capt-of}
    \usepackage{booktabs,array,tabularx}
    \usepackage[table]{xcolor}
    \usepackage{pifont}
    \usepackage{url}
    \makeatletter
    \newcommand{\kiwiteaser}[1]{\gdef\@kiwiteaser{#1}}
    \makeatother
    \newtoks\affiliationnote
    \newtoks\correspondingauthor
    \usepackage{balance}
\fi

\usepackage{graphicx}
\usepackage{amsmath}
\usepackage{amssymb}
\usepackage{stfloats}

\newcommand{\egocam}{GO~3}
\newcommand{\methodshort}{KIWI}

\newcommand{\projecturl}{https://lingfeng.moe/KIWI}
\newcommand{\projectlink}{\expandafter\url\expandafter{\projecturl}}

\title{\LARGE \bf Kinematic Interface for the Wild: Modular Bimanual Loco-Manipulation Capture from 360$^{\circ}$ Cameras Alone}

\author{
Benjamin~C.~Yang*,
Weiying~Wang*,
Shenggao~Li,
Keming~Yan,
Sasha~Wilkinson,
Zelin~Wang,
Yip~Fun~Yeung,
Lingfeng~Sun$^\dag$
}

\renewcommand{\correspondingmark}{$^{\dag}$}
\affiliationnote={*Equal contribution. All authors are with Autel US.}
\correspondingauthor={Lingfeng Sun, \href{mailto:lingfengsun1996@gmail.com}{lingfengsun1996@gmail.com}}
\paperurl={\projecturl}

\kiwiteaser{%
  \begin{center}
    \includegraphics[width=0.98\textwidth]{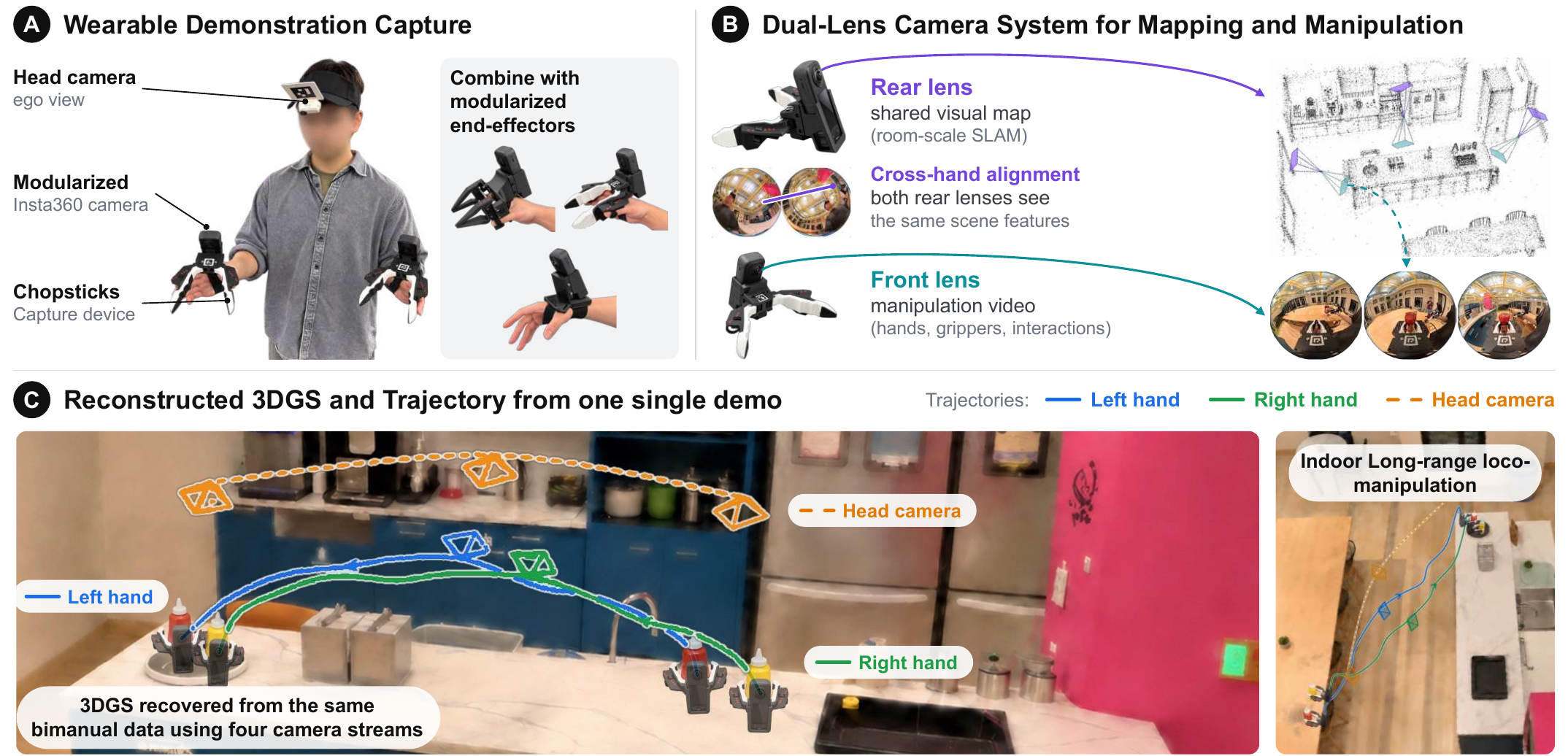}
    \captionof{figure}{\textbf{\methodshort{} overview.} Camera-only wearable capture of bimanual demonstrations. (A)~A head camera and a modular dual-lens camera on each hand, compatible with interchangeable end-effectors. (B)~Rear lenses build a shared room-scale map and align the two hands through common scene features; front lenses record the manipulation. (C)~One bimanual demo yields a 3DGS scene plus 6-DoF left-hand, right-hand, and head trajectories in a single frame, from countertop tasks to room-scale loco-manipulation.}
    \label{fig:overview}
  \end{center}%
}

\begin{document}

\begin{abstract}
A wrist-mounted camera for UMI-style data collection must do two jobs: record the manipulation and localize in the scene.
Most handheld devices localize online from workspace-facing views crowded by hands and objects, or add dedicated tracking hardware.
Room-scale bimanual capture therefore still tends to instrument the operator or the scene for accurate localization.
We present KIWI (Kinematic Interface for the Wild), a capture kit whose only electronics are off-the-shelf cameras. Our core system splits the two jobs across the two lenses of a 360$^{\circ}$ camera.
The rear lens faces the room and builds a shared metric map that registers both hands, and an optional head camera, in one frame without workspace co-visibility; the front lens records the manipulation, and offline IMU fusion bridges front-lens tracking loss.
Through our quick-release plate, the camera module attaches to chopstick grippers, parallel-jaw grippers, hand-wrist mounts, or robot flanges.
Across six bimanual recordings, combining the rear and front lenses failed to localize only 0.1\% of query frames, whereas front-only bimanual feature alignment failed on 24.8\% of frames and lost one recording entirely; against evaluation fiducials, localization error stayed within 4.5\,mm.
KIWI's recovered poses were sufficiently consistent for the four wrist streams alone to reconstruct the scene as a 3D Gaussian splat.
Hardware and software will be fully open-sourced on our \href{\projecturl}{website}.
\end{abstract}

\ifcustomclass
  \maketitle
  \clearpage
\else
  \makeatletter\IEEEaftertitletext{\vspace{-0.9\baselineskip}\@kiwiteaser}\makeatother
  \maketitle
  \thispagestyle{empty}\pagestyle{empty}
\fi

\section{Introduction}
\label{sec:introduction}

\begin{table*}[tp]
  \caption{Demonstration interfaces: collection hardware and recovered state.}
  \label{tab:system-comparison}
  \centering
  \fontsize{8}{9.2}\selectfont
  \definecolor{KiwiGreen}{HTML}{42651D}
  \definecolor{KiwiInk}{HTML}{35412B}
  \definecolor{KiwiTint}{HTML}{F0F5E7}
  \definecolor{KiwiLabel}{HTML}{F7F8F3}
  \definecolor{KiwiHeader}{HTML}{E4ECD8}
  \definecolor{KiwiRule}{HTML}{CCD5BE}
  \definecolor{KiwiAmber}{HTML}{876215}
  \definecolor{KiwiBrown}{HTML}{684A3A}
  \definecolor{KiwiMute}{HTML}{8B927D}
  \newcommand{\ch}[2]{{\fontsize{7.5}{8.6}\selectfont\textcolor{KiwiGreen}{\textbf{#1}} \cite{#2}}}
  \newcommand{\cmk}{\textcolor{KiwiGreen}{\ding{51}}}
  \newcommand{\xmk}{\textcolor{KiwiBrown}{\ding{55}}}
  \newcommand{\cmpnr}{\textcolor{KiwiAmber}{\textbf{NR}}}
  \newcommand{\cmpna}{\textcolor{KiwiMute}{N/A}}
  \color{KiwiInk}
  \arrayrulecolor{KiwiRule}
  \setlength{\arrayrulewidth}{0.5pt}
  \setlength{\extrarowheight}{1.5pt}
  \setlength{\tabcolsep}{2.5pt}
  \renewcommand{\arraystretch}{1.12}
  \begin{tabularx}{\textwidth}{@{}>{\columncolor{KiwiLabel}[0pt][\tabcolsep]\raggedright\arraybackslash\bfseries}p{0.95in}
      !{\color{KiwiGreen}\vrule width 0.6pt}
      >{\columncolor{KiwiTint}\centering\arraybackslash}X
      !{\color{KiwiGreen}\vrule width 0.6pt}
      *{10}{>{\columncolor{white}\centering\arraybackslash}X}
      >{\columncolor{white}[\tabcolsep][0pt]\centering\arraybackslash}X@{}}
    \hline
    \rowcolor{KiwiHeader}
    \textcolor{KiwiGreen}{Feature} & \textcolor{KiwiGreen}{\textbf{KIWI} (ours)} & \ch{UMI}{chi2024umi} & \ch{UMI-3D}{wang2026umi3d} & \ch{iPhUMI}{patel2026iphumi} & \ch{FastUMI}{zhaxizhuoma2025fastumi} & \ch{RDT2}{liu2026rdt2} & \ch{YUBI}{ohkawa2026yubi} & \ch{EgoMI}{yu2025egomi} & \ch{HiFi-UMI}{wei2026hifiumi} & \ch{XRZero-G0}{wang2026xrzero} & \ch{DAS Fingers}{genrobot2026dasfingers} \\
    \hline
    Sensor suite & Insta360 X5, \egocam & GoPro Hero9 & Livox Mid-360, global-shutter cam & iPhone 15~Pro & GoPro Hero9, RealSense T265 & Global-shutter cam, Vive Tracker & ELP fisheye, Quest~3S, rotary encoder & Quest~3S, ZED~2i, wrist cams & Pro\-pri\-etary & PICO~4, ego + wrist cams & Pro\-pri\-etary \\ \hline
    Extra compute free & \cmk & \cmk & \xmk & \cmk & \xmk & \xmk & \xmk & \xmk & \xmk & \xmk & \xmk \\ \hline
    Wrist 6-DoF trajectory & Offline SLAM & Online SLAM & Online LiDAR + IMU & Online ARKit & Online RealSense T265 & Online Vive IR & Online VR & Online VR & Offline SLAM + marker & Online VR & Online VIO \\ \hline
    Ego view & \cmk & \xmk & \xmk & \cmk & \xmk & \xmk & \cmk & \cmk & \cmk & \cmk & \cmk \\ \hline
    Inter-hand pose & \cmk~map & \cmk~map & \cmpna & \cmk~ARKit & \cmpnr & \cmk~Vive & \cmk~VR & \cmk~VR & \cmk~head & \cmk~VR & \cmk~Ego \\ \hline
    Scene coverage & $360^\circ$ wrist + ego & $155^\circ$ + mirrors & $185^\circ$ + LiDAR & Head + wrist & $169^\circ$ wrist & Wrist & $180^\circ$ + chest & Head + wrist & $\sim 200^\circ$ + head & Head + wrist & $150^\circ$ + Ego \\ \hline
    Floor + height & \cmk & \cmpnr & \cmpnr & \cmpnr & \cmpnr & \cmpnr & \cmpnr & \cmpnr & \cmpnr & \cmpnr & \cmpnr \\ \hline
    Open source design & \cmk & \cmk & \cmk & \cmk & \cmk & \xmk & \cmk & \xmk & \xmk & \xmk & \xmk \\ \hline
  \end{tabularx}
  \arrayrulecolor{black}
  \par\vspace{3pt}\noindent
  \begin{minipage}{\textwidth}
    \scriptsize
    \cmk{} yes; \xmk{} no; NR: not reported or not verified; N/A: not applicable.
    Extra compute free means the in-hand tool carries no electronics or compute beyond the recording camera.
    Online trajectory estimation runs incrementally (device tracking or odometry, even when applied to recordings); offline estimation optimizes over the completed recording.
    Open source design \cmk{} means public build files and software sufficient to reproduce the interface, excluding commercial sensor firmware and tracking services; partial releases (code, data, or SDKs only) count as \xmk{}, and KIWI's \cmk{} reflects its planned release.  \end{minipage}
\end{table*}

Robot learning from human demonstrations needs more than video: both hands' tool poses in a common coordinate frame, inter-hand distance, gripper state, and time-aligned camera views.
Tool trajectories in world coordinates provide targets for robot replay and full-body inverse kinematics (IK), while real-to-sim transfer increasingly calls for a model of the scene itself~\cite{abouchakra2026realissim}.
Wearable capture systems obtain these signals in three ways, each with a cost.
Tracker-based devices take tool pose from a VR headset and controller or from a tracking module on the tool~\cite{xu2025exumi,ohkawa2026yubi,yu2025egomi,zhaxizhuoma2025fastumi}, and recording their wrist cameras adds a computer on the tool or carried by the operator~\cite{xu2025exumi,ohkawa2026yubi}.
Phone-based rigs let one device record and track online~\cite{shafiullah2023dobbe}, but remain costly and rely on proprietary tracking software.
Camera-only handheld grippers~\cite{chi2024umi} remove the extra hardware, but the workspace-facing wrist camera must then localize from a view crowded by hands and objects, and registering the two hands relies on features both views share; exUMI~\cite{xu2025exumi} reports that less than 60\% of vanilla UMI recordings survive processing.
Adding a scene reconstruction to the dataset usually requires a separate, dedicated scan.

KIWI (Kinematic Interface for the Wild) provides full state estimation and scene reconstruction from off-the-shelf cameras alone by splitting the work across the two lenses of a $360^\circ$ camera (Insta360 X5) on each wrist (Fig.~\ref{fig:overview}).
The front lens records the manipulation.
The rear lens faces the room and builds one shared metric, gravity-aligned map per scene; both wrist cameras localize in it, so the two hands are registered in one frame without being required to view the same workspace. An optional head camera is posed from a marker on the operator's hat that the rear lenses observe.
An offline-optimized per-wrist factor graph then fuses each camera's IMU with front and rear lens localizations, so tracking survives when the view is occluded or too close to the scene.
The remainder of the KIWI stack also makes use of the camera-native recordings.
The cameras are synchronized from shared ambient audio, the gripper opening is read from the front-facing video, and a plane estimate anchors the dataset to a ground or tabletop surface, enabling replay verification of loco-manipulation demonstrations and supplying the floor reference that loco-manipulation controllers depend on~\cite{zhu2025relic}.
Finally, all four wrist streams are utilized to reconstruct the scene as a 3D Gaussian splat (3DGS), frame-aligned with the gripper trajectories.

Our quick-release interface makes the hardware modular: the same calibrated camera module moves between handheld tools and their robot-mounted counterparts.
Adding a tool requires a matching adapter and a camera-to-tip transform.
We support our ergonomic two-finger \textit{Chopstick} gripper, an ALOHA-style parallel-jaw gripper~\cite{zhao2023aloha}, a bare-hand wrist mount, and matching robot flanges.

We evaluate the pipeline on six bimanual scenarios.
Cross-hand front-view localization, the camera-only alternative to a rear map, leaves 24.8\% of the 246{,}116 query frames without visual support and collapses on one recording, whereas per-hand front maps placed in the shared rear reference support more than 99.9\%.
Against fixed fiducial markers used only for evaluation, the median translation error is 4.3 and 4.5\,mm for the two hands, and the four wrist streams of a single recording reconstruct the scene as a 3DGS that renders held-out views. Our contributions are:
\begin{itemize}
  \item \textbf{Two-lens state estimation.} The rear lens builds a shared map from the demonstrations themselves and registers both hands, and optionally the head camera, without workspace co-visibility; offline fusion with the front lens and IMU bridges front-view loss. Evidence: a same-recording comparison with cross-hand front-view localization, and swapping which wrist builds the map.
  \item \textbf{Complete loco-manipulation episodes from cameras alone.} Both hands' 6-DoF tool poses and their height above the floor, gripper opening, synchronized views, and the head-marker pose, together with a 3D Gaussian splat of the scene recovered without a separate scan, all from the cameras' own video, IMU, and audio.
  \item \textbf{One sensing core, an open design.} KIWI's open-source hardware and software let users build on the provided suite of grippers and robot-side mounts to create custom end effectors and robot adapters, reusing the calibrated camera module and reconstruction pipeline.
\end{itemize}

\section{Related Work}
\label{sec:related-work}

\begin{figure}[t]
  \centering
  \includegraphics[width=\columnwidth]{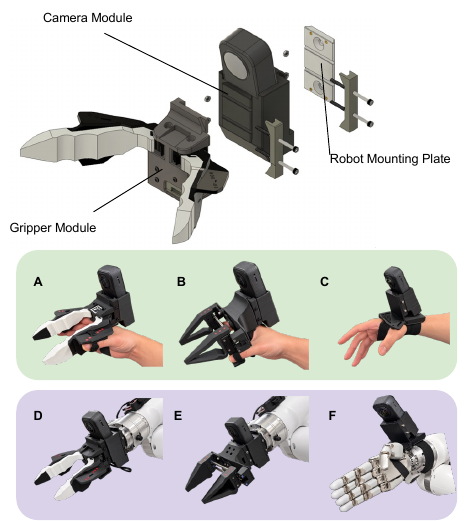}
  \caption{\textbf{Modular hardware overview.}
    Top: exploded view of the three-part design: an interchangeable
    gripper module, the shared camera module, and a robot mounting plate.
    Bottom: (A) chopstick and (B) parallel-jaw handheld grippers and
    (C) the bare-hand wrist cuff; (D, E, F) the robot-mounted twins.}
  \label{fig:hardware-overview}
\end{figure}


\subsection{Handheld Interfaces for Manipulation Data Collection}

Handheld gripper interfaces collect manipulation demonstrations without a robot by mounting cameras on the tool and recovering actions offline.
UMI~\cite{chi2024umi} established this pattern: a printed parallel-jaw gripper under a wrist fisheye camera, monocular-inertial ORB-SLAM3 trajectory recovery against a per-scene map, jaw width tracked from finger fiducials, and a bimanual mode that registers both grippers through the shared map while pairing their frames to within $1/60$\,s.
However, the recovery chain's reliability has been the recurring pain point---exUMI~\cite{xu2025exumi} reports that less than 60\% of vanilla UMI recordings survive processing---and successor systems responded by instrumenting the tool: rotary encoders and 6-DoF trackers~\cite{xu2025exumi}.
YUBI's~\cite{ohkawa2026yubi} bimanual workflow features finger-driven yielding jaws with magnetic aperture encoders, a wrist camera per tool, and VR-headset 6-DoF tracking, supporting a stationary desk rig as well as a portable chest-mounted mode, evaluated in a matched usability study against UMI.
Phone-based rigs such as iPhUMI~\cite{patel2026iphumi} draw view, depth, and online pose from a single retail phone, but remain expensive with limited post-processing capabilities.
KIWI takes the opposite tradeoff from the instrumented branch: the tool stays fully passive, with no encoders, batteries, or wiring, and every measurement, including jaw state, comes from the wrist cameras' own recordings of video, inertia, and audio, paid for by the offline reconstruction of Sec.~\ref{sec:reconstruction}.

\subsection{Egocentric Sensing for Manipulation Capture}

Egocentric demonstrations are attracting increasing interest as a scalable source of data for robot learning.
EgoDex~\cite{hoque2026egodex} and EgoMimic~\cite{kareer2025egomimic} record bare-hand manipulation, leaving an embodiment gap between human hands and robot grippers that must be addressed during transfer.
EgoMI~\cite{yu2025egomi} reduces this mismatch by collecting demonstrations with robot grippers and synchronized head and hand tracking, but relies on a VR headset and wired electric grippers.
HiFi-UMI~\cite{wei2026hifiumi} localizes gripper-mounted marker cubes from a head-mounted stereo rig, making hand tracking dependent on the markers remaining visible to the head cameras.
Table~\ref{tab:system-comparison} summarizes collection hardware and recovered state across representative handheld and wearable interfaces.

\section{Hardware and Capture Design}
\label{sec:hardware}

Two requirements shape the kit (Fig.~\ref{fig:hardware-overview}).
\emph{Nothing on the human operator but the cameras}: no electronics on the tool, no worn device, no carried computer, no dangling wires, and no base stations, so a session is two cameras and printed parts.

\emph{One shared quick-release}: a common interface connects the camera module to each handheld tool and its robot-side counterpart.
The camera module and its mounting base form a reusable sensing core, allowing the kit's embodiment to change rapidly as task requirements change.

\subsection{Camera Module and Shared Quick-Release}
\label{sec:hardware-camera-mount}

\textit{Camera module.}
The camera module is the only sensor on the tool: a retail Insta360 X5 secured in a 3D-printed holder with a quick-release connector for attachment to each tool.
Its two fisheye lenses, IMU, and microphone record to the camera's own storage, and what each lens is for is the subject of Sec.~\ref{sec:reconstruction}; nothing on the module is wired to anything else.
Each physical unit is calibrated once and carries a dated artifact set: per-lens fisheye intrinsics, a camera--IMU extrinsic and time shift per lens, Allan-deviation noise densities, and an audio-to-video delay per recording mode.
Two camera holder designs are provided: a robust frame that securely joins the robot flange to the gripper, and a lightweight, ergonomic version for human use without any connection to the robot.

\textit{Quick-release system.}
The quick-release system is a modified Arca-Swiss plate; as the camera industry's most common plate standard, it keeps compatible holders for other cameras easy to source or design.
Two modifications adapt the standard plate: the locking screw moves to the center of the dovetail, avoiding breaks along printed layer bonds and saving space, and the safety stop screw becomes a precise positioning feature, fixing a known geometry between the camera and each tool tip and between the robot flange and the camera module so that recorded camera motion converts to tool-tip motion without a robot during capture.

\subsection{Handheld End Effectors}
\label{sec:hardware-end-effectors}

Pistol-grip interfaces such as UMI~\cite{chi2024umi} offset the fingers from the pinch point.
This geometry reduces contact feedback, encouraging excessive force and hindering fine grasp adjustments~\cite{ohkawa2026yubi}.
KIWI instead uses yielding, finger-driven jaws that follow the natural pinch and preserve mechanical feedback.

\begin{figure}[t]
  \centering
  \includegraphics[width=0.85\columnwidth]{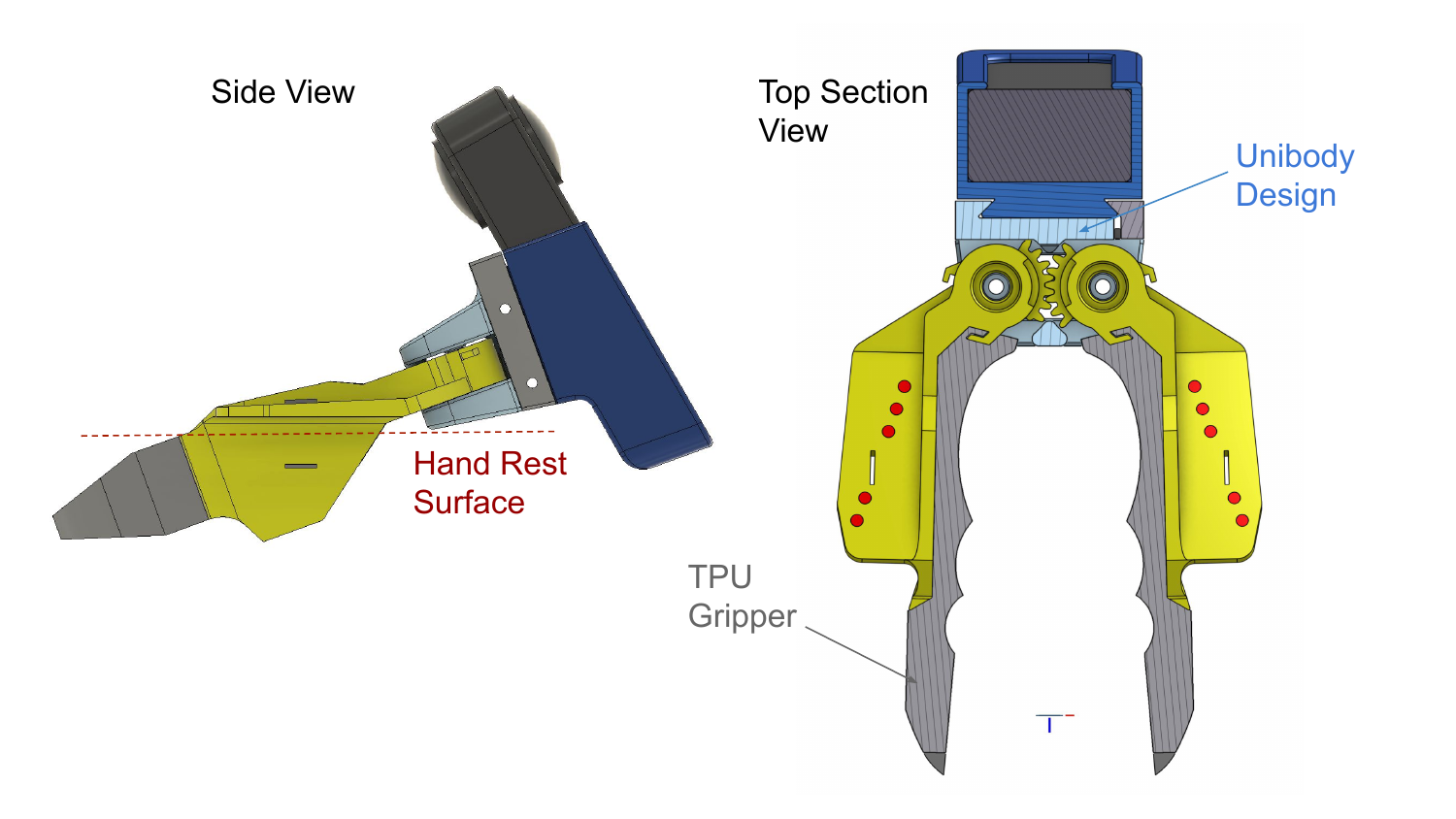}
  \caption{\textbf{Chopstick gripper geometry and hinge mechanism.}
    Left: side view of the assembled gripper and camera, showing the angled grip that provides clearance for the camera body.
    Right: section view of the bearing-supported jaw pivots and meshing gears.}
  \label{fig:chopstick-design}
\end{figure}

\textit{Chopstick gripper.}
The chopstick gripper implements this principle as a passive angle gripper, sharing the hinged-jaw design of YUBI~\cite{ohkawa2026yubi}, Generalist's UMI-style gripper~\cite{generalist2026web}, and XRZero-G0's G-shaped gripper~\cite{wang2026xrzero}.
Two printed jaws pivot on integrated bearings to reduce friction (Fig.~\ref{fig:chopstick-design}), and a rubber band reopens them on release.
The chopstick opening at the tip spans 195\,mm maximum.

The main design constraint was camera placement.
A centered view of the gripper's contact area helps limit lens distortion, but the Insta360 X5's body is longer than those of webcam, RealSense, or ZED camera modules and can interfere with the operator's hand.
Inspired by Generalist's design, we angle the grip to provide clearance for the camera body (Fig.~\ref{fig:chopstick-design}, left).
During tasks with the jaws pointing downward, both friction between the fingers and the angled grip and an elastic palm strap help support the device, reducing the holding force required from the fingers.

The brim on each jaw helps keep the operator's hand out of the camera view and carries a row of red dots (Fig.~\ref{fig:chopstick-design}, right) from which reconstruction recovers the hinge angle.

The chopstick module weighs 367\,g including the camera.
For comparison, YUBI's handheld unit with its controller weighs 319\,g, while UMI weighs approximately 780\,g~\cite{ohkawa2026yubi}.

\textit{Parallel-jaw gripper.}
The parallel-jaw gripper follows the ALOHA form factor~\cite{zhao2023aloha} and reuses many design elements from the chopstick gripper.
The design retains ALOHA's double rails but removes the surrounding support structure for hand clearance.

\textit{Hand module.}
The wrist cuff carries the camera module on the quick-release plate and shares the state-estimation and scene-reconstruction pipeline.

\subsection{Optional Head Module}
\label{sec:hardware-head-module}

The optional head module rigidly mounts an Insta360 \egocam{} and an ArUco tag on a hat (Fig.~\ref{fig:overview}(A)).
The compact, lightweight camera and tag share a printed mount.

\subsection{Chopstick Design for 3D Printing}
\label{sec:hardware-fabrication}

The FDM-printed chopstick gripper uses a unibody to reduce joints, with bearing seats and screw tension designed to limit hinge play while preserving smooth, finger-driven jaw motion.
Parameterized bearing-seat allowances accommodate different printing processes and materials while preserving the shared mounting interface.

\subsection{Capture Protocol and Cost}
\label{sec:hardware-protocol-cost}

Each demonstration, or episode, begins and ends with Insta360's built-in voice commands ``start recording'' and ``stop recording.''
The operator performs tabletop or walking tasks; recordings are stored onboard and offloaded afterward.
The two X5 cameras are the major cost at US\$1{,}099.98 (store.insta360.com, 2026-09), and the optional \egocam{} adds US\$349.99 (2026-08).

\section{Offline Demonstration Reconstruction}
\label{sec:reconstruction}

\begin{figure*}[t]
  \centering
  \includegraphics[width=\textwidth]{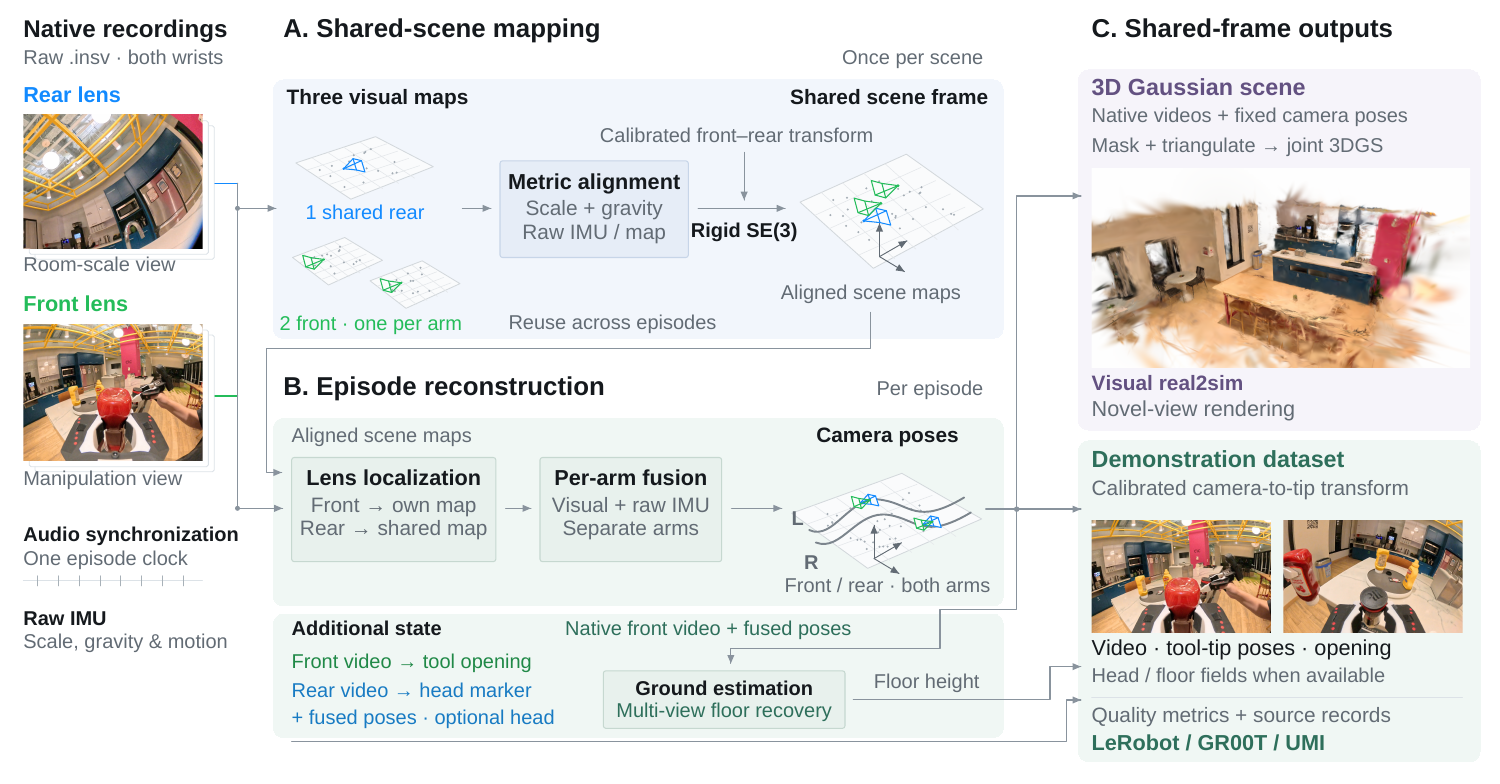}
  \caption{\textbf{KIWI offline pipeline.} (A) One shared rear map and two front maps recover metric scale and gravity from IMU measurements and are aligned using calibrated front--rear geometry. (B) Each episode reuses these maps for independent per-wrist visual--inertial pose estimation, complemented by tool-opening, head-marker, and ground estimation from
  native videos and recovered poses. (C) Outputs include synchronized demonstration datasets with tool-tip poses and quality records, alongside a 3D Gaussian scene reconstructed from native fisheye videos and aligned camera poses for visual real-to-sim.}
  \label{fig:pipeline}
\end{figure*}

\subsection{Time Synchronization}
\label{sec:time-sync}

KIWI aligns the independently recorded wrist and ego cameras on a common timeline using their microphones alone, with no simultaneous starts and no timecode hardware.
For each camera pair, spectral fingerprints~\cite{wang2003audio,bryan2012multicamera} vote for a coarse audio lag and GCC-PHAT~\cite{knapp1976gcc} refines it on the waveforms; a match is accepted only with strong, unambiguous support.
A sound heard at local audio times $a_i$ and $a_j$ gives the measured lag $\Delta_{ij}=a_i-a_j$.
Correcting it by each camera's internal A/V delay $d_i$ aligns the videos:
\begin{equation}
    q_j-q_i=\Delta_{ij}+d_j-d_i,
    \label{eq:sync-av}
\end{equation}
where $q_i$ places stream $i$ on the shared timeline.
For matching camera models and modes a verified common-delay assumption cancels the $d$ terms, and redundant pairs must close consistently around triangles.
Because a short shared sound proves an offset but not clock stability, a robust fit across the overlap classifies each pair as \emph{offset-only} or \emph{offset-and-drift} evidence, and well-supported drift violations cause rejection.
Residual A/V bias, acoustic propagation, and unobserved clock changes bound the final accuracy.

\subsection{Shared-Scene State Estimation}
\label{sec:state}

\begin{figure}[!h]
  \centering
  \includegraphics[width=0.9\columnwidth]{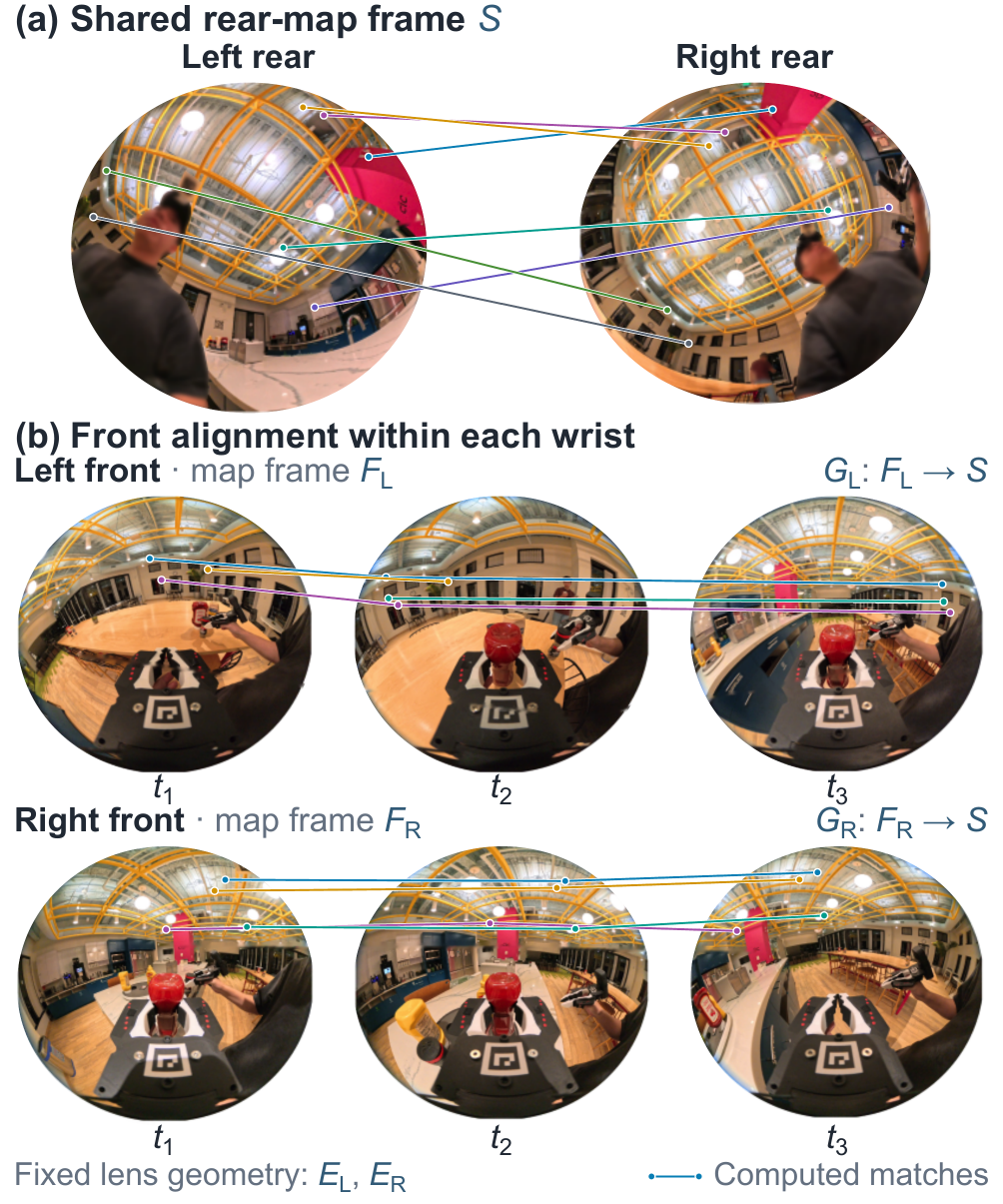}
  \caption{\textbf{Shared-scene registration from fisheye views.}
    (a) Rear-view correspondences illustrate the common scene reference $S$.
    (b) Temporal correspondences in each front stream support separate map
    frames $F_L$ and $F_R$.
    Fixed front--rear calibration $E_a$ and synchronized localizations
    determine the rigid placements $G_a:F_a\rightarrow S$
    in Eq.~\eqref{eq:scene-placement}.
    Colored lines indicate computed image matches.}
  \label{fig:fisheye-registration}
\end{figure}

We construct a reusable scene reference from a randomly designated bimanual demonstration and use it to reconstruct the remaining demonstrations in that environment.
Vision-only ORB-SLAM3~\cite{campos2021orbslam3} builds a front-view map for each wrist and one shared rear-view map from a designated wrist's rear stream.
The front maps retain each wrist's manipulation-view observations; the rear map supplies an environmental reference that both wrists subsequently query.
Each stream uses its calibrated fisheye projection~\cite{kannala2006generic}, with hand/tool regions masked during front-map construction~\cite{ronneberger2015unet}.
The SLAM engine estimates visual camera poses and scene structure; metric recovery and cross-map registration follow separately.

We exploit the offline nature of our demonstration reconstruction to separate visual map construction from metric recovery.
UMI~\cite{chi2024umi} also processes recorded demonstrations, but uses an incremental monocular visual--inertial SLAM pipeline that continues optimizing after initialization from a saved map.
In KIWI, metric recovery is a separate fit over the completed visual maps, using valid relocalizations and corresponding raw IMU measurements across the full mapping recording, even when a shorter interval was used to build the maps.
This allows motion occurring later in an ordinary demonstration to contribute to the metric reference reused by subsequent demonstrations.
With visual poses and camera--IMU calibration fixed, a robust visual--inertial fit estimates each map's scale, gravity direction, and accelerometer bias~\cite{campos2020inertial}.
The recovered scale converts translations to meters, and the gravity-alignment rotation defines a $z$-up frame for each map.

The metric rear-map frame defines the common scene frame $S$ (Fig.~\ref{fig:fisheye-registration}).
For wrist $a\in\{L,R\}$, let $F_a$ be its metric front-map frame and let $C_f,C_r,I$ denote that wrist's front camera, rear camera, and IMU frames.
We use $T_{AB}$ to map coordinates from $B$ to $A$.
The two camera--IMU calibrations~\cite{furgale2013kalibr} give a fixed front-to-rear transform $E_a$; synchronized front and rear localizations of the mapping demonstration determine a constant map placement $G_a$:
\begin{equation}
\begin{aligned}
E_a&=T_{C_rI}T_{C_fI}^{-1},\\
G_aT_{F_aC_f}(t)&\approx T_{SC_r}(t)E_a,
\qquad G_a\in\mathrm{SE}(3).
\end{aligned}
\label{eq:scene-placement}
\end{equation}
Robust fitting estimates only $G_a$, keeping the recovered scales and physical lens calibration fixed.
The fixed lens geometry connects the two views of each device even when they observe different scene regions.
Because both wrists' rear streams localize in the same rear map, the two placements $G_L,G_R$ already establish bimanual co-registration; no additional left--right trajectory registration is performed.
We freeze these placements and the metric maps for all subsequent demonstrations.

For each demonstration, each front stream queries its own front map and both rear streams query the shared rear map without updating map structure.
Queries use unmasked frames with image geometry matched to the map; camera and IMU timestamps use the synchronization described in Sec.~\ref{sec:time-sync}.
Each successful front or rear localization is converted into a body-pose observation in $S$ using the frozen placement and camera--IMU calibration:
\begin{equation}
\begin{aligned}
Z_a^f(t)&=G_a\widetilde T_{F_aC_f}(t)T_{C_fI},\\
Z_a^r(t)&=\widetilde T_{SC_r}(t)T_{C_rI}.
\end{aligned}
\label{eq:state-observations}
\end{equation}
Here the tilde denotes a queried camera pose in its metric map frame.
These constructions give both lenses a common spatial meaning before temporal fusion; failed localizations remain missing observations.

A separate graph for each wrist estimates $\mathcal X_a=\{X_{a,i},\mathbf v_{a,i},\mathbf b_{a,i}\}$ at that wrist's native front-camera timestamps, where $X_{a,i}=T_{SI}(t_i)$ is body pose, $\mathbf v_{a,i}$ is scene-frame velocity, and $\mathbf b_{a,i}$ contains accelerometer and gyroscope biases.
The front and rear observations in Eq.~\eqref{eq:state-observations} provide unary pose factors; raw IMU preintegration~\cite{forster2017preintegration} jointly connects pose, velocity, and bias at adjacent times.
The graph minimizes
\begin{equation}
\begin{aligned}
\mathcal X_a^*&=\underset{\mathcal X_a}{\operatorname{argmin}}
\left(\mathcal E_a^{I}+\mathcal E_a^{f}+\mathcal E_a^{r}+\mathcal E_a^{0}\right),\\
\mathcal E_a^c&=\sum_{i\in\mathcal I_{a,c}}
\rho_\delta\!\left(\left\|
\operatorname{Log}_{\mathrm{SE}(3)}\!\left((Z_{a,i}^c)^{-1}X_{a,i}\right)
\right\|_{\Sigma_c}\right).
\end{aligned}
\label{eq:state-objective}
\end{equation}
For $c\in\{f,r\}$, $Z_{a,i}^c$ is the body observation associated with node $i$, $\mathcal I_{a,c}$ is the admitted observation set, and $\rho_\delta$ applies Huber loss to the covariance-weighted pose-residual norm, with $\|\mathbf r\|_\Sigma^2=\mathbf r^\top\Sigma^{-1}\mathbf r$.
The combined inertial term $\mathcal E_a^I$ includes bias evolution; $\mathcal E_a^0$ contains the initial velocity and zero-centered bias priors.
Front observations use their original timestamps; rear body poses may be interpolated only across short, contiguous valid tracking intervals.
Visual factors are temporally thinned to limit their aggregate weight while keeping every front frame as a state node.
The scene reference, calibration, and timestamps remain fixed, and each wrist's motion is estimated in a separate graph conditioned on this reference.
We solve the graph with Levenberg--Marquardt in GTSAM~\cite{dellaert2017factor} and output $T_{SC_f,i}=X_{a,i}^*T_{C_fI}^{-1}$.

\subsection{Auxiliary State Estimation}
\subsubsection{End-Effector State Estimation}
\label{sec:eef}

\begin{figure}[t]
  \centering
  \includegraphics[width=\columnwidth]{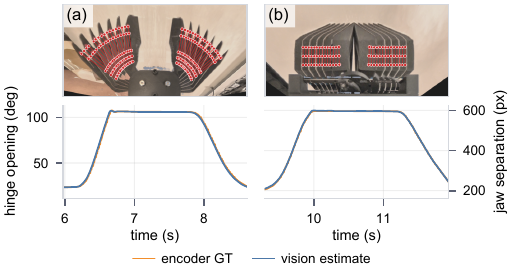}
  \caption{\textbf{End-effector state for (a) the chopstick and (b) the
    parallel-jaw gripper.}
    Top: fronto-parallel views of the gripper opening with detected dots
    and per-jaw line fits at uniform opening steps.
    The end-effector module estimates state independently on every frame of the
    120\,fps recordings.
    Bottom: single-cycle excerpts of the vision estimate (blue) versus
    the servo-encoder ground truth (orange).
    RMSE against the encoder is $0.17^\circ$ in (a) and $1.9$\,px in
    (b).}
  \label{fig:eef-dot-trails}
\end{figure}

KIWI recovers the gripper's state per frame directly from the wrist video. For the chopstick gripper, each jaw carries a row of printed red dots in a plane rigidly attached to the tool body, and a body-mounted ArUco tag~\cite{garrido2014aruco}, used once during calibration, anchors that plane in the camera frame. Each rectified frame maps to the marker frame: a fronto-parallel view of the gripper dots (Fig.~\ref{fig:eef-dot-trails}).
A line is fit to each side's extracted dots, providing the opening between the two jaw lines.


\subsubsection{Optional Head Pose}
\label{sec:head}
An optional head-mounted ArUco marker~\cite{garrido2014aruco} of known size is observed in either rear view.
IPPE~\cite{collins2014ippe} applied to fisheye-undistorted corners gives $T^a_{C_rM}(t)$ after planar-ambiguity rejection, where $M$ denotes the marker frame.
From the optimized front-camera output and $E_a$ in Eq.~\eqref{eq:scene-placement}, we obtain rear poses $T^a_{SC_r,i}=T^a_{SC_f,i}E_a^{-1}$, then interpolate them to detection times.
Each detection becomes a scene-frame marker observation
\begin{equation}
\widetilde T^a_{SM}(t)=T^a_{SC_r}(t)T^a_{C_rM}(t).
\label{eq:head-observation}
\end{equation}
An ego-camera optical trajectory additionally requires marker-to-camera calibration.

\subsubsection{Optional Ground Reference}
\label{sec:ground}

\begin{figure}[!htbp]
  \centering
  \includegraphics[width=\columnwidth]{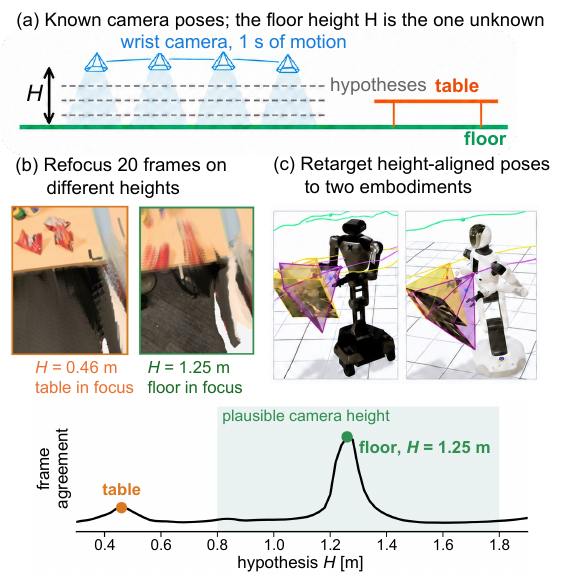}
  \caption{\textbf{Ground-height estimation and pose retargeting.}
(a) Candidate horizontal planes from known camera poses.
(b) Agreement across 20 refocused frames selects the floor
($H=1.25$\,m) within the plausible camera-height range;
the table lies at $H=0.46$\,m.
(c) Height-aligned poses retargeted to two embodiments.
$H$ denotes camera-to-plane distance.}
  \label{fig:ground-estimation}
\end{figure}

Using the metric, gravity-aligned camera trajectories from state estimation, we estimate one horizontal floor per episode.
Let $z_f$ be its scene-frame height and $\mathbf c^f_{a,i}$ the translation of $T^a_{SC_f,i}$.
With $\mathbf e_3=(0,0,1)^\top$, the floor and camera height are
\begin{equation}
\begin{aligned}
\Pi_f(z_f)&=\{\mathbf p:\mathbf e_3^\top\mathbf p=z_f\},\\
h_{a,i}&=\mathbf e_3^\top\mathbf c^f_{a,i}-z_f.
\end{aligned}
\label{eq:ground-reference}
\end{equation}
We estimate $z_f$ from short front-video windows with downward views and horizontal camera translation.

Two complementary channels propose surface heights: photometric plane sweeping~\cite{collins1996plane} compares views warped onto candidate horizontal planes (Fig.~\ref{fig:ground-estimation}), while rectified motion stereo~\cite{hirschmuller2008stereo} reconstructs points from frame pairs of the same camera.
Each selects the lowest sufficiently supported surface within a configured camera-to-floor height band.
Agreeing estimates are combined; a single accepted channel can also supply a measurement, while conflicting accepted estimates reject the window.
After temporal outlier rejection, a weighted median pools accepted heights from both wrists into the episode floor level.

\providecommand{\kiwievaldir}{notes/state-estimation-review-20260915/evaluation-v3}
\section{Evaluation}
\label{sec:evaluation}

\subsection{Pipeline Performance}

\begin{figure}[t]
  \centering
  \includegraphics[width=\columnwidth]{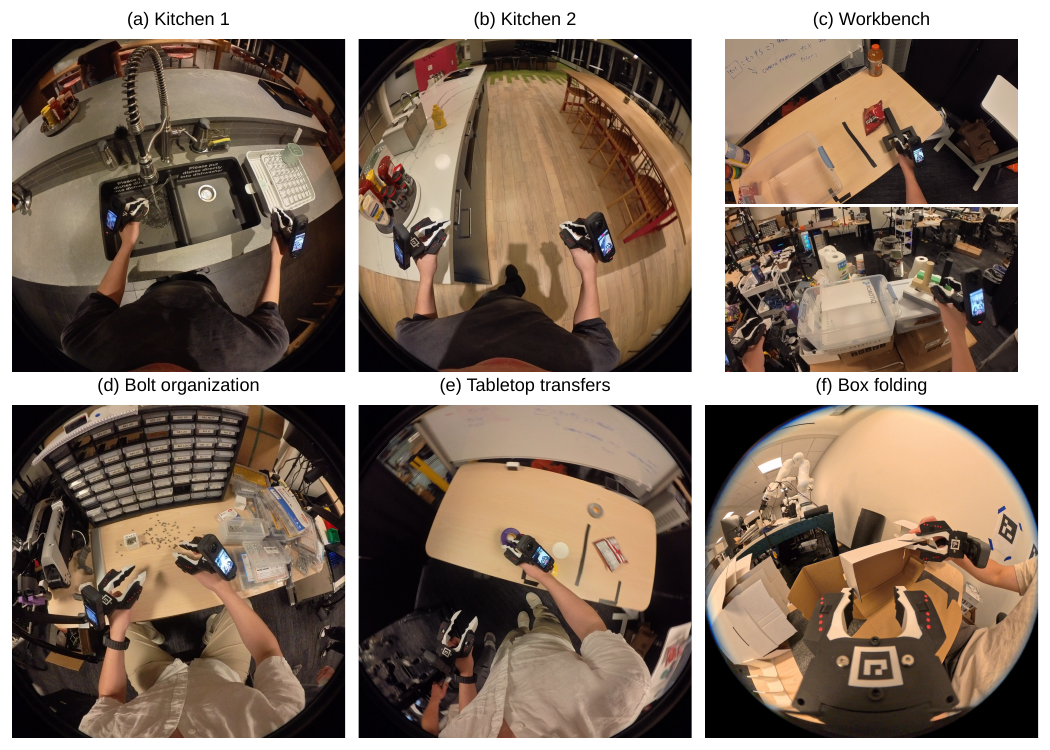}
  \caption{\textbf{Task and workspace diversity across six configurations.}
  Images are auxiliary egocentric views and the wrist-front view of
  box folding.}
  \label{fig:eval-scenes}
\end{figure}

We evaluate KIWI's pose-estimation pipeline across six recording configurations: Kitchen 1 and Kitchen 2, manipulation with operator movement around a small workbench, bimanual bolt organization, tabletop object transfers, and large-box folding (Fig.~\ref{fig:eval-scenes}). Each evaluated recording also supplies its scene maps.

Pose estimation uses $960\times960$ images at 119.88\,Hz from all four wrist-camera streams.
On an AMD Ryzen 9 9950X, four-stream localization of the 331.1\,s Workbench recording takes 353.2\,s (median of two runs), or 1.07 times the recording duration.
This timing measures localization from prepared images and scene maps.

\subsection{Front-Only Ablation}

We compare visual tracking availability in the full pipeline with a front-only configuration. The front-only baseline localizes the right-front camera in the left-front map. Across the six bimanual recordings, cross-hand front-view localization fails on 24.8\% of 246116 query frames. KIWI's per-hand front maps and shared rear reference provide native visual support on more than 99.9\% of the same frames (Table~\ref{tab:eval-map-sharing}). A frame is supported when a valid front observation or a rear observation is available at its timestamp; rear observations may be interpolated within a continuous tracked run over at most 50\,ms. The rear map establishes the common scene frame, allowing each front camera to maintain tracking against its own map without requiring continuous cross-hand feature overlap.

\begin{table}[t]
  \centering
  \footnotesize
  \setlength{\tabcolsep}{2pt}
  \caption{Visual tracking availability.}
  \label{tab:eval-map-sharing}
  \begin{tabular}{@{}lrrrr@{}}
    \toprule
    Recording & Duration & Frames & Front-only (\%) & KIWI (\%) \\
    \midrule
    Kitchen 1 & 317.0 & 38,006 & 95.509 & 99.997 \\
    Kitchen 2 & 294.8 & 35,340 & 99.997 & 99.997 \\
    Workbench & 331.1 & 39,692 & 93.734 & 99.997 \\
    Bolt organization & 414.5 & 49,687 & 0.000 & 99.998 \\
    Tabletop transfers & 338.4 & 40,571 & 91.765 & 99.998 \\
    Box folding & 357.2 & 42,820 & 91.240 & 99.998 \\
    \bottomrule
  \end{tabular}
  \par\smallskip
  \begin{minipage}{\columnwidth}
    \footnotesize Front-only and KIWI are evaluated on the same right-front query frames across six bimanual recordings.
  \end{minipage}
\end{table}

\subsection{Synchronization Accuracy}

We evaluate audio-based synchronization against an optical ground truth that shares no mechanism with the audio path: every camera films the same monitor displaying a rolling QR clock, each camera's frame timestamps are robustly fit to the displayed clock, and differencing two fits pins the true cross-camera offset to 0.6--2.4\,ms standard error. On a held-out capture, the estimator recovers the offset of a same-model X5 pair from audio alone to within 1.9\,ms of the optical truth. For the mixed-model pair (X5 wrist camera vs.\ GO 3 ego camera), after applying the per-mode A/V constant ($d$ in Eq.~\eqref{eq:sync-av}), the recovered offsets land within 2.5 and 4.7\,ms of the optical truth for the two X5 bodies, and the two resulting placements of the ego camera agree to 0.3\,ms. All residuals are sub-frame for every stream, including the X5s at 119.88\,Hz.

\subsection{Marker-Referenced Pose Accuracy}

We place two fixed fiducial markers in the workspace to evaluate tracking accuracy and compare the estimated wrist-camera trajectories with marker-derived reference poses. The markers have a nominal side length of 58\,mm and provide eight-corner PnP reference poses from calibrated images. Their planar arrangement is fitted separately for each wrist using only marker observations from the first 40\% of the sequence, then held fixed. A fixed-scale rigid alignment is fitted on the first 40\% of the sequence, followed by a 10\% gap; the final 50\% is used for scoring. Independent per-hand alignments give median translation ATEs of 4.31 and 4.48\,mm for the left and right hands. Applying the same left-fitted alignment to both hands gives 4.31 and 6.92\,mm, respectively, retaining their estimated relative placement. These measurements use 17965 left and 18142 right poses with valid marker references.

\subsection{Qualitative Scene Reconstruction}

\begin{figure}[t]
  \centering
  \includegraphics[width=\columnwidth]{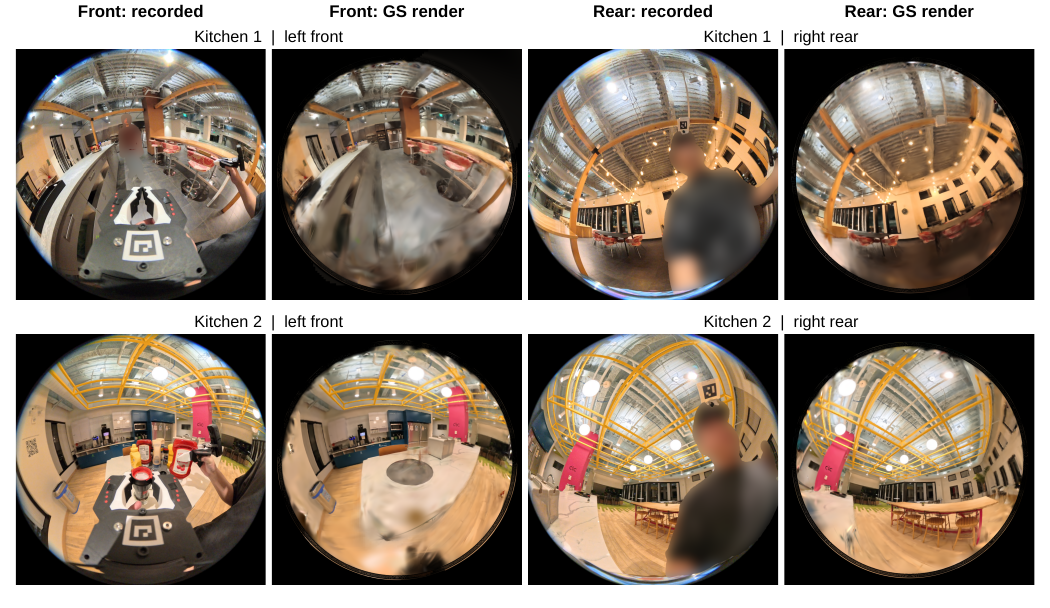}
  \caption{\textbf{Held-out 3DGS renders.} Recorded images (left) and held-out 3DGS renders (right) from Kitchen 1 and Kitchen 2. Camera poses are fixed; rear-view people are blurred.}
  \label{fig:eval-gs}
\end{figure}

To assess whether the recovered poses support a shared visual scene, we reconstruct Kitchen 1 and Kitchen 2 using earlier saved KIWI trajectories. We sample native $1920\times1920$ fisheye images at approximately 1\,Hz, associate them with the recovered camera poses at their exposure times, and mask people, capture hardware, and selected manipulated objects. Using fixed camera poses, calibration, and synchronization, we triangulate training-view feature matches to initialize one Gaussian scene and fit it jointly to all four wrist-camera streams using 3DGUT~\cite{wu20253dgut} for 30000 optimization steps. Every eighth temporal group is withheld across all streams from Gaussian fitting.
Renders preserve room layout and recognizable static structures in held-out images (Fig.~\ref{fig:eval-gs}), providing qualitative evidence of multi-view pose consistency.

\balance

\end{document}